\documentclass{llncs}

\usepackage{times}

\usepackage{amssymb}
\usepackage{amsmath}

\usepackage{amsthm}
\usepackage{balance}
\usepackage{color}
\usepackage{enumerate}
\usepackage{enumitem}
\usepackage{epstopdf}
\usepackage{etoolbox}
\usepackage{float}
\usepackage{graphicx}
\usepackage{hyperref}
\usepackage{multirow}
\usepackage{url}
\usepackage{xspace}
\usepackage{wrapfig}
\usepackage{rotating}
\usepackage{comment}

\hypersetup{
    colorlinks,
    linkcolor={red},
    citecolor={black},
    urlcolor={black}
}

\newcounter{myctr}

\newenvironment{myitemize}{\begin{list}{$\bullet$}
{\setlength{\topsep}{1mm}\setlength{\itemsep}{0.25mm}
\setlength{\parsep}{0.1mm}
\setlength{\itemindent}{0mm}\setlength{\partopsep}{0mm}
\setlength{\labelwidth}{15mm}
\setlength{\leftmargin}{4mm}}}{\end{list}}

\newcommand{\hide}[1]{}

\newcommand{\Real}{{\mathbb R}}

\DeclareMathOperator{\bfx}{{\bf x}}
\DeclareMathOperator{\bfd}{{\bf d}}

\DeclareMathOperator{\bfdelta}{{\bf \delta}}
\DeclareMathOperator{\bfM}{{\bf M}}

\DeclareMathOperator{\calC}{{\mathcal C}}

\DeclareMathOperator{\bbR}{{\mathbb R}}

\DeclareMathOperator*{\signf}{sign} 

\DeclareMathOperator*{\KL}{KL}

\newcommand{\FGSM}{{\rm FGSM}}

\newcommand{\VAT}{{\rm VAT}}

\newcommand{\reals}{\mathbb{R}}
\newcommand{\bools}{\mathbb{B}}

\newcommand{\pred}{\sigma}
\newcommand{\U}[1]{\mathtt{U}_{#1}}

\newcommand{\F}[1]{\mathtt{F}_{#1}}
\newcommand{\G}[1]{\mathtt{G}_{#1}}
\newcommand{\rob}[3]{\rho(#1,#2,#3)}

\title{Semantic Adversarial Deep Learning}
\author{Tommaso Dreossi$^{1}$, Somesh Jha$^{2}$, \and Sanjit A. Seshia$^{1}$}
\institute{University of California at Berkeley, Berkeley$^{1}$\\
University of Wisconsin, Madison$^{2}$\\ 
\email{dreossi@berkeley.edu, jha@cs.wisc.edu, sseshia@berkeley.edu}}

\begin{document}

\maketitle

\begin{abstract}
Fueled by massive amounts of data, models produced by machine-learning
(ML) algorithms, especially deep neural networks, are being used 
in diverse domains where
trustworthiness is a concern, including automotive systems, finance,
health care, natural language processing, and malware detection. 
Of particular concern is the use of ML
algorithms in cyber-physical systems (CPS), such as self-driving cars and
aviation, where an adversary can cause serious consequences.

However, existing approaches to generating adversarial examples and devising
robust ML algorithms mostly ignore the {\em semantics} and {\em context} of the overall
system containing the ML component. For example, in an autonomous vehicle
using deep learning for perception, not every adversarial example 
for the neural network might lead to a harmful consequence. Moreover,
one may want to prioritize the search for adversarial examples towards those
that significantly modify the desired semantics of the overall system. 
Along the same lines, existing algorithms for constructing robust ML 
algorithms ignore the specification of the overall system. 
In this paper, we argue that the semantics and specification of the 
overall system has a crucial role to play in this line of research. 
We present preliminary research results that support this claim.
\end{abstract}

\section{Introduction}
\label{sec:intro}

{\it Machine learning (ML)} algorithms, fueled by massive amounts of data, are
increasingly being utilized in several domains, including healthcare, finance, and
transportation. 
Models produced by ML algorithms, especially {\em deep neural networks} (DNNs), 
are being deployed in domains where trustworthiness is a big
concern, such as automotive systems~\cite{NVIDIATegra},
finance~\cite{PayPal}, health care~\cite{alipanahi2015predicting},
computer vision~\cite{krizhevsky2012imagenet}, speech
recognition~\cite{hinton2012deep}, natural language
processing~\cite{pennington2014glove}, and
cyber-security~\cite{dahl2013large,shin2015recognizing}. Of particular
concern is the use of ML (including deep learning)
in {\it cyber-physical systems} (CPS)~\cite{leeseshia-16}, where the
presence of an adversary can cause serious consequences.  For example,
much of the technology behind autonomous and driver-less vehicle
development is ``powered'' by machine
learning~\cite{btd+16,iwdriverless,self-driving-2016}. DNNs
have also been used in airborne collision
avoidance systems for unmanned aircraft (ACAS Xu)~\cite{DASC:2016}. 
However, {\it in designing and deploying these
algorithms in critical cyber-physical systems, the presence of an
active adversary is often ignored.}

{\em Adversarial machine learning (AML)} is a field concerned with the analysis
of ML algorithms to adversarial attacks, and the use of such analysis in making
ML algorithms robust to attacks. It is part of the broader agenda
for safe and verified ML-based systems~\cite{russell2015letter,seshia-arxiv16}. 
In this paper, we first give a brief survey
of the field of AML, with a particular focus on deep learning.
We focus mainly on attacks on outputs or models that are
produced by ML algorithms that occur {\it after training} or
``external attacks'', which are especially relevant to cyber-physical
systems (e.g., for a driverless car the ML algorithm used for
navigation has been already trained by the manufacturer once the ``car
is on the road''). These attacks are more realistic and are distinct
from other type of attacks on ML models, such as attacks that poison
the training data (see the paper~\cite{huang2011adversarial} for a
survey of such attacks).
We survey attacks caused by \emph{adversarial examples}, which are 
inputs crafted by adding small, often imperceptible, perturbations to 
force a trained ML model to misclassify.  
%
%

We contend that the work on adversarial ML, while important and useful, is
not enough. In particular, we advocate for the increased use of
{\em semantics} in adversarial analysis and design of ML algorithms.
{\em Semantic adversarial learning} explores a space of semantic
modifications to the data, uses system-level semantic specifications
in the analysis, utilizes semantic adversarial examples in training,
and produces not just output labels but also additional semantic
information. Focusing on deep learning, we explore these ideas
and provide initial experimental data to support them.

\noindent
{\bf Roadmap.} Section~\ref{sec:background} provides the relevant
background.  A brief survey of adversarial analysis is given 
in Section~\ref{sec:attacks}. Our proposal for semantic adversarial
learning is given in Section~\ref{sec:semantic-analysis}.

\section{Background}
\label{sec:background}
\subsubsection*{Background on Machine Learning}

Next we describe some general concepts in machine learning (ML).  We
will consider the supervised learning setting.  Consider a sample
space $Z$ of the form $X \times Y$, and an ordered training set $S \;
= \; ((x_i,y_i))_{i=1}^m$ ($x_i$ is the data and $y_i$ is the
corresponding label).  Let $H$ be a hypothesis space (e.g., weights
corresponding to a logistic-regression model).  There is a loss
function $\ell: H \times Z \mapsto \Real$ so that given a hypothesis
$w \in H$ and a sample $(x,y) \in Z$, we obtain a loss $\ell(w, (x,y))$.
We consider the case where we want to minimize the loss over the
training set $S$,
\[ L_S(w) = \frac{1}{m}\sum_{i=1}^m \ell(w, (x_i, y_i)) \; + \; \lambda \mathcal{R}(w).  \]
In the equation given above, $\lambda >0$ and the term $\mathcal{R}(w)$ is called the
{\it regularizer} and enforces ``simplicity'' in $w$.  Since $S$ is
fixed, we sometimes denote $\ell_i(w) = \ell(w, (x_i, y_i))$ as a
function only of $w$. We wish to find a $w$ that minimizes $L_S (w)$ or
we wish to solve the following optimization problem:
\[
\min_{w \in H} L_S (w)
\]

\noindent
{\bf Example:} We will consider the example of logistic regression.
In this case $X = \Real^n$, $Y = \{ +1 , -1 \}$, $H = \Real^n$, and
the loss function $\ell (w, (x,y))$ is as follows ($\cdot$ represents
the dot product of two vectors):
\[
\log \left( 1 + e^{-y (w^T \cdot x )} \right)
\]
If we use the $L_2$ regularizer (i.e. $\mathcal{R}(w) = \parallel w \parallel_2$),
then $L_S (w)$ becomes:
\[
\frac{1}{m}\sum_{i=1}^m \log \left( 1 + e^{-y_i (w^T \cdot x_i )} \right) \; + \; 
\lambda  \parallel w \parallel_2
\]

\noindent\textbf{Stochastic Gradient Descent.}  {\it Stochastic
  Gradient Descent (SGD)} is a popular method for solving optimization
tasks (such as the optimization problem $\min_{w \in H} L_S (w)$ we
considered before).  In a nutshell, SGD performs a series of updates
where each update is a gradient descent update with respect to a small
set of points sampled from the training set.  Specifically, suppose
that we perform SGD $T$ times.  There are two typical forms of SGD: in
the first form, which we call {\sf Sample-SGD}, we uniformly and randomly
sample $i_t \sim [m]$ at time $t$, and perform a gradient descent
based on the $i_t$-th sample $(x_{i_t}, y_{i_t})$:
\begin{align}
  w_{t+1} = G_{\ell_t, \eta_t}(w_t) = w_t - \eta_t\ell_{i_t}'(w_t)
\label{eqn:sgd-update}
\end{align}
where $w_t$ is the hypothesis at time $t$, $\eta_t$ is a
parameter called the {\it learning rate}, and $\ell_{i_t}' (w_t)$
denotes the derivative of $\ell_{i_t} (w)$ evaluated at $w_t$.  We
will denote $G_{\ell_t, \eta_t}$ as $G_t$.  In the second form, which
we call {\sf Perm-SGD}, we first perform a random permutation of $S$,
and then apply Equation~\ref{eqn:sgd-update} $T$ times by cycling
through $S$ according to the order of the permutation.  The process of
SGD can be summarized as a diagram:
\begin{align*}
  w_0
  \mathop{\longrightarrow}^{G_1}
  w_1
  \mathop{\longrightarrow}^{G_2}
  \cdots
  \mathop{\longrightarrow}^{G_t}
  w_t
  \mathop{\longrightarrow}^{G_{t+1}}
  \cdots
  \mathop{\longrightarrow}^{G_T}
  w_T
\end{align*}


\noindent\textbf{Classifiers.}  
The output of the learning algorithm gives us a {\it classifier},
which is a function from $\Re^n$ to $\calC$, where $\Re$ denotes the
set of reals and $\calC$ is the set of class labels.  To emphasize
that a classifier depends on a hypothesis $w \in H$, which is the
output of the learning algorithm described earlier, we will write it
as $F_w$ (if $w$ is clear from the context, we will sometimes simply
write $F$).  For example, after training in the case of logistic
regression we obtain a function from $\Re^n$ to $\{ -1,
+1 \}$. Vectors will be denoted in boldface, and the $r$-th component
of a vector $\bfx$ is denoted by $\bfx[r]$.  

Throughout the paper, we refer to the function $s(F_w)$ as the {\it
softmax layer} corresponding to the classifier $F_w$.  In the case of
logistic regression, $s(F_w) (\bf x)$ is the following tuple (the
first element is the probability of $-1$ and the second one is the
probability of $+1$):
\[
\langle \frac{1}{1 + e^{w^T \cdot \bfx}} , \frac{1}{1 + e^{-w^T \cdot \bfx}} \rangle
\]
Formally, let $c = | \calC |$ and $F_w$ be a classifier, we let $s(F_w)$ be
the function that maps $\bbR^n$ to $\bbR_+^c$ such that $\|s(F_w)(\bfx)\|_1
= 1$ for any $\bfx$ (i.e., $s(F_w)$ computes a probability vector).  We
denote $s(F_w)(\bfx)[l]$ to be the probability of $s(F_w)(\bfx)$ at label
$l$. 
Recall that the softmax function from
$\Real^k$ to a probability distribution over $\{ 1,\cdots, k \} \;
= \; [k]$ such that the probability of $j \in [k]$ for a vector
$\bfx \in \Real^k$ is
\[
\frac{ e^{\bfx[j]}}{\sum_{r=1}^k e^{\bfx[r]}}
\]

Some classifiers $F_w (\bfx)$ are of the form $\arg\max_{l} s(F_w)
(\bfx) [l]$ (i.e., the classifier $F_w$ outputs the label with the
maximum probability according to the ``softmax layer''). For example,
in several deep-neural network (DNN) architectures the last layer is
the {\it softmax} layer. 
We are assuming that the reader is a familiar with basics of
deep-neural networks (DNNs). For readers not familiar with DNNs we can
refer to the excellent book by Goodfellow, Bengio, and
Courville~\cite{Goodfellow-et-al-2016}.

\subsubsection*{Background on Logic}

Temporal logics are commonly used for specifying desired and undesired
properties of systems. For cyber-physical systems, it is common to
use temporal logics that can specify properties of real-valued signals
over real time, such as {\em signal temporal logic} (STL)~\cite{maler2004monitoring} 
or {\em metric temporal logic} (MTL)~\cite{koymans-90-mtl}.

A \emph{signal} is a function $s : D \to S$, with $D \subseteq \reals_{\geq 0}$ 
an interval and either $S \subseteq \bools$ or $S \subseteq \reals$,
where $\bools = \{\top, \bot\}$ and $\reals$ is the set of reals. 
Signals defined on $\bools$ are called \emph{booleans}, while
those on $\reals$ are said \emph{real-valued}. 
A \emph{trace} $w = \{s_1,\dots, s_n\}$ is a finite set of real-valued signals
defined over the same interval $D$. We use variables $x_i$ to denote
the value of a real-valued signal at a particular time instant.

Let $\Sigma = \{\pred_1, \dots, \pred_k \}$ be a finite set of predicates $\pred_i : \reals^n \to \bools$,
with $\pred_i \equiv p_i(x_1, \dots, x_n) \lhd 0$, $\lhd \in \{ <, \leq \}$, and $p_i : \reals^n \to \reals$
a function in the variables $x_1, \dots, x_n$.
An STL formula is defined by the following grammar:
\begin{equation}
	\varphi := \pred \, | \, \neg \varphi \, | \, 
                    \varphi \wedge \varphi \, | \, \varphi\ \U{I}\ \varphi
\end{equation}
where $\pred \in \Sigma$ is a predicate and $I \subset \reals_{\geq 0}$ is a closed
non-singular interval. Other common temporal operators
can be defined as syntactic abbreviations in the usual way, like for instance
$\varphi_1 \vee \varphi_2 := \neg ( \neg \varphi_1 \wedge \varphi_2 )$,
$\F{I}\ \varphi := \top\ \U{I}\ \varphi$, or $\G{I}\ \varphi\ := \neg \F{I}\ \neg \varphi$.
Given a $t \in \reals_{\geq 0}$, a shifted interval $I$
is defined as $t + I = \{t + t' \mid t' \in I\}$.
The qualitative (or Boolean) semantics of STL is given in the
usual way:
\begin{definition}[Qualitative semantics]
	Let $w$ be a trace, $t \in \reals_{\geq 0}$, and $\varphi$ be an STL formula.
	The \emph{qualitative semantics} of $\varphi$ is inductively defined as follows:
	\begin{equation}
		\begin{split}
			w,t \models  \pred \text{ iff } 			&\pred(w(t)) \text{ is true} \\
			w,t \models \neg\varphi \text{ iff } 		& w,t \not\models \varphi \\
			w,t \models \varphi_1 \wedge \varphi_2 \text{ iff }	& w,t \models \varphi_1 \text{ and } w,t \models \varphi_2 \\
			w,t \models \varphi_1 \U{I} \varphi_2 \text{ iff }	& \exists t' \in t + I \text{ s.t. } w,t' \models \varphi_2 \text{ and } \forall t'' \in [t,t'], w,t'' \models \varphi_1 \\
		\end{split}
	\end{equation}
\end{definition}
A trace $w$ satisfies a formula $\varphi$ if and only if $w,0 \models \varphi$, in short $w \models \varphi$.
STL also admits a quantitative or robust semantics, which we
omit for brevity. This provides quantitative information on 
the formula, telling how strongly the specification is satisfied or violated for a given trace.

\hide{
\begin{definition}[Quantitative/Robust Semantics]\label{def:rob}
	Let $w$ be a trace, $t \in \reals_{\geq 0}$, and $\varphi$ be an STL formula.
	The \emph{robustness} $\rho$ of $\varphi$ for a trace $w$ at time $t$ is defined as:
	\begin{equation}
		\begin{split}
			\rob{p(x_1, \dots, x_n) \lhd 0}{w}{t} = &\ p(w(t)) \text{ with } \lhd \in \{<,\leq \} \\
			\rob{\neg\varphi}{w}{t} = &\ -\rob{\varphi}{w}{t} \\
			\rob{\varphi_1 \wedge \varphi_2}{w}{t} = &\ \min( \rob{\varphi_1}{w}{t}, \rob{\varphi_2}{w}{t} ) \\
			\rob{\varphi_1 \U{I} \varphi_2}{w}{t} = &\ \sup_{t' \in t+I} \min( \rob{\varphi_2}{w}{t'}, \inf_{t''[t,t']} \rob{\varphi_1}{w}{t''} ) \\
		\end{split}
	\end{equation}
\end{definition}

Note that a trace $w$ satisfies a formula $\varphi$ if and only if $\rob{\varphi}{w}{0} > 0$.

Differently from classic qualitative semantics, STL robustness
provides quantitative information on the evaluated formula, i.e., it tells how strongly the specification is satisfied or violated
by the considered trace.
}

\section{Attacks}
\label{sec:attacks}

There are several types of attacks on ML algorithms.  For excellent
material on various attacks on ML algorithms we refer the reader
to~\cite{huang2011adversarial,barreno2010security}.  For example, in
{\it training time} attacks an adversary wishes to poison a data set
so that a ``bad'' hypothesis is learned by an ML-algorithm.  This
attack can be modeled as a game between the algorithm $ML$ and an
adversary $A$ as follows:
\begin{itemize}
\item $ML$ picks an ordered training set 
$S \; = \; ((x_i,y_i))_{i=1}^m$.
\item $A$ picks an ordered training set 
$\widehat{S} \; = \; ((\hat{x_i},\hat{y_i}))_{i=1}^r$, where
$r$ is $\lfloor \epsilon m \rfloor$.
\item $ML$ learns on $S \cup \widehat{S}$ by essentially minimizing
\[
\min_{w \in H} L_{ S \cup \widehat{S}} (w) \; .
\]
\end{itemize}
The attacker wants to maximize the above quantity and thus chooses $
\widehat{S}$ such that \\ $\min_{w \in H} L_{ S \cup \widehat{S}} (w)$ is
maximized. For a recent paper on certified defenses for such attacks
we refer the reader to~\cite{steinhardt:2017}. In {\it model
  extraction} attacks an adversary with black-box access to a
classifier, but no prior knowledge of the parameters of a ML algorithm
or training data, aims to duplicate the functionality of (i.e., “steal”)
the classifier by querying it on well chosen data points. For
an example, model-extraction attacks see~\cite{model-extraction}.

In this paper, we consider {\it test-time attacks}. We assume that the
classifier $F_w$ has been trained without any interference from the
attacker (i.e. no training time attacks). Roughly speaking, an
attacker has an image $\bfx$ (e.g. an image of stop sign) and wants to
craft a perturbation $\bfdelta$ so that the label of $\bfx+\bfdelta$
is what the attacker desires (e.g. yield sign). The next sub-section 
describes test-time attacks in detail. We will sometimes refer to $F_w$
as simply $F$, but the hypothesis $w$ is lurking in the background (i.e.,
whenever we refer to $w$, it corresponds to the classifier $F$).

\subsection{Test-time Attacks}

The adversarial goal is to take any input vector $\bfx \in \Re^n$ and
produce a minimally altered version of $\bfx$, \emph{adversarial
  sample} denoted by $\bfx^\star$, that has the property of being
misclassified by a classifier $F : \Re^n \rightarrow \calC$.
Formally speaking, an adversary wishes to solve the following
optimization problem:

\[
\begin{array}{lr}
\min_{\bfdelta \in \Re^n}  & \mu ( \bfdelta ) \\
\mbox{such that} & F( \bfx + \bfdelta ) \in T \\
& \bfdelta \cdot \bfM \; = \; 0
\end{array}
\]

The various terms in the formulation are $\mu$ is a metric on $\Re^n$,
$T \subseteq \calC$ is a subset of the labels (the reader should think
of $T$ as the target labels for the attacker), and $\bfM$ (called the
{\it mask}) is a $n$-dimensional $0-1$ vector of size $n$.  The
objective function minimizes the metric $\mu$ on the perturbation
$\bfdelta$.  Next we describe various constraints in the formulation.
 
\begin{itemize}
\item $F( \bfx + \bfdelta ) \in T$\\ The set $T$ constrains the
  perturbed vector $\bfx+\bfdelta$~\footnote{The vectors are added
    component wise} to have the label (according to $F$) in the set
  $T$. For {\it mis-classification} problems the label of $\bfx$ and
  $\bfx+\bfdelta$  are different, so we have $T = \calC - \{ F(\bfx)
  \}$. For {\it targeted mis-classification} we have $T = \{ t \}$
  (for $t \in \calC$), where $t$ is the target that an attacker wants
  (e.g., the attacker wants $t$ to correspond to a yield sign).

\item $\bfdelta \cdot \bfM \; = \; 0$\\ The vector $M$ can be
  considered as a mask (i.e., an attacker can only perturb a dimension
  $i$ if $M [i] \; = \; 0$), i.e., if $M [i] \; = \; 1$ then $\bfdelta
  [i]$ is forced to be $0$.  Essentially the attacker can only perturb
  dimension $i$ if the $i$-th component of $M$ is $0$, which means
  that $\delta$ lies in $k$-dimensional space where $k$ is the number
  of non-zero entries in $\Delta$. This constraint is important if an
attacker wants to target a certain area of the image (e.g., glasses of
in a picture of person) to perturb.

\item {\it Convexity} \\ Notice that even if the metric $\mu$ is
  convex (e.g., $\mu$ is the $L_2$ norm), because of the constraint
  involving $F$, the optimization problem is {\it not convex} (the
  constraint $\bfdelta \cdot \bfM \; = \; 0$ is convex).  In general,
  solving convex optimization problems is more tractable non-convex
  optimization~\cite{Wright:book}.
\end{itemize}

Note that the constraint $\bfdelta \cdot \bfM \; = \; 0$ essentially
constrains the vector to be in a lower-dimensional space and does 
add additional complexity to the optimization problem. Therefore, for
the rest of the section we will ignore that constraint and work with
the following formulation:

\[
\begin{array}{lr}
\min_{\bfdelta \in \Re^n}  & \mu ( \bfdelta ) \\
\mbox{such that} & F( \bfx + \bfdelta ) \in T 
\end{array}
\]

\noindent\textbf{FGSM mis-classification attack -} This
algorithm is also known as the \emph{fast gradient sign method
  (FGSM)}~\cite{goodfellow2014explaining}.  The adversary crafts an
adversarial sample $\bfx^\star = \bfx+\bfdelta$ for a given legitimate
sample $\bfx$ by computing the following perturbation:
\begin{equation}
\label{eq:goodfellow-perturbation}
\bfdelta  = \varepsilon \signf(\nabla_{\bfx} L_F (\bfx))
\end{equation}
The function $L_F (\bfx)$ is a shorthand for $\ell (w,\bfx,l(\bfx))$,
where $w$ is the hypothesis corresponding to the classifier $F$,
$\bfx$ is the data point and $l(\bfx)$ is the label of $\bfx$
(essentially we evaluate the loss function at the hypothesis
corresponding to the classifier).  The gradient of the function $L_F$
is computed with respect to $\bfx$ using sample $\bfx$ and label $y =
l(\bfx)$ as inputs. Note that $\nabla_{\bfx} L_F (\bfx)$ is an
$n$-dimensional vector and $\signf(\nabla_{\bfx} L_F (\bfx))$ is a
$n$-dimensional vector whose $i$-th element is the sign of the $
\nabla_{\bfx} L_F (\bfx)) [i]$.  The value of the \emph{input
  variation parameter} $\varepsilon$ factoring the sign matrix
controls the perturbation's amplitude. Increasing its value increases
the likelihood of $\bfx^\star$ being misclassified by the classifier
$F$ but on the contrary makes adversarial samples easier to detect by
humans. The key idea is that FGSM takes a step {\it in the direction
  of the gradient of the loss function} and thus tries to maximize
it. Recall that SGD takes a step in the direction that is opposite to
the gradient of the loss function because it is trying to minimize the
loss function.

\noindent\textbf{JSMA targeted mis-classification attack
  -} This algorithm is suitable for targeted
misclassification~\cite{papernot2015limitations}. We refer to this
attack as JSMA throughout the rest of the paper.  To craft the
perturbation $\bfdelta$, components are sorted by decreasing
{\it adversarial saliency value}. The adversarial saliency value
$S(\bfx,t)[i]$ of component $i$ for an adversarial target class $t$ is
defined as:
\begin{equation}
\label{eq:saliency-map-increasing-features}
S(\bfx,t)[i] = \left\lbrace
\begin{array}{c}
0  \mbox{ if }   \frac{\partial s(F) [t] (\bfx)}{\partial \bfx [i]}<0  \mbox{ or } \sum_{j\neq t} \frac{\partial s(F)[j] (\bfx)}{\partial \bfx [i]} > 0\\
 \frac{\partial s(F)[t] (\bfx)}{\partial \bfx [i]} \left| \sum_{j\neq t}  
\frac{\partial s(F)[j] (\bfx)}{\partial \bfx [i]} \right| \mbox{ otherwise}
\end{array}\right.
\end{equation}
where matrix $J_F=\left[\frac{\partial s(F)[j] (\bfx)}{\partial \bfx
    [i]}\right]_{ij}$ is the Jacobian matrix for the output of the
softmax layer $s(F)(\bfx)$. Since $\sum_{k \in \calC }  s(F)[k] (\bfx) \; = \; 1$,
we have the following equation:
\begin{eqnarray*}
 \frac{\partial s(F) [t] (\bfx)}{\partial \bfx [i]} &
= & -\sum_{ j \neq t} \frac{\partial s(F) [j] (\bfx) }{\partial \bfx [i]} 
\end{eqnarray*}
The first case corresponds to the scenario if
changing the $i$-th component of $\bfx$ takes us further away from the
target label $t$. Intuitively, $S(\bfx,t)[i]$ indicates how likely is
changing the $i$-th component of $\bfx$ going to ``move towards'' the
target label $t$. Input components $i$ are added to perturbation
$\bfdelta$ in order of decreasing adversarial saliency value
$S(\bfx,t)[i]$ until the resulting adversarial sample $\bfx^\star =
\bfx + \bfdelta$ achieves the target label $t$. The perturbation
introduced for each selected input component can vary. Greater
individual variations tend to reduce the number of components
perturbed to achieve misclassification.

\noindent\textbf{CW targeted mis-classification attack.}  The
CW-attack~\cite{Carlini:2017} is widely believed to be one of the most
``powerful'' attacks. The reason is that CW cast their problem as an
unconstrained optimization problem, and then use state-of-the art
solver (i.e. Adam~\cite{Adam}). In other words, they leverage the
advances in optimization for the purposes of generating adversarial
examples.

In their paper Carlini-Wagner consider a
wide variety of formulations, but we present the one that performs best
according to their evaluation. The optimization problem corresponding
to CW is as follows:
\[
\begin{array}{lr}
\min_{\bfdelta \in \Re^n}  & \mu ( \bfdelta ) \\
\mbox{such that} & F( \bfx + \bfdelta ) \; = \; t
\end{array}
\]
CW use an existing solver (Adam~\cite{Adam}) and thus need to make
sure that each component of $\bfx + \bfdelta$ is between $0$ and $1$
(i.e.  valid pixel values). Note that the other methods did not face
this issue because they control the ``internals''  of the algorithm (i.e.,
CW used a solver in a ``black box'' manner). We introduce a new vector
${\bf w}$ whose $i$-th component is defined according to the following
equation:
\begin{eqnarray*}
\bfdelta [i] & = & \frac{1}{2} (\tanh ({\bf w}[i]) + 1) - \bfx [i]
\end{eqnarray*} 
Since $-1 \leq \tanh({\bf w}[i]) \leq 1$, it follows that 
$0 \leq \bfx [i] + \bfdelta [i] \leq 1$. In terms of this new variable the
optimization problem becomes:
\[
\begin{array}{lr}
\min_{{\bf w} \in \Re^n}  & \mu ( \frac{1}{2} (\tanh ({\bf w}) + 1) - \bfx ) \\
\mbox{such that} & F( \frac{1}{2} (\tanh ({\bf w}) + 1))  \; = \; t
\end{array}
\]

Next they approximate the constraint ($F( \bf{x}) \; = \; t$)
with the following function:
\begin{eqnarray*}
g(\bf{x}) & = & \max \left( \max_{i \not= t }  Z(F)(\bfx)[i] \; - \; 
Z(F)(\bfx)[t], -\kappa \right)
\end{eqnarray*}
In the equation given above $Z(F)$ is the input of the DNN to the softmax layer
(i.e $s(F)(\bfx) = \mbox{softmax}(Z(F)(\bfx))$) and $\kappa$ is a confidence
parameter (higher $\kappa$ encourages the solver to find adversarial examples
with higher confidence). The new optimization formulation is as follows:
\[
\begin{array}{lr}
\min_{{\bf w} \in \Re^n}  & \mu ( \frac{1}{2} (\tanh ({\bf w}) + 1) - \bfx ) \\
\mbox{such that} & g( \frac{1}{2} (\tanh ({\bf w}) + 1))  \;\leq \; 0
\end{array}
\]
Next we incorporate the constraint into the objective function as follows:
\[
\begin{array}{lr}
\min_{{\bf w} \in \Re^n}  & \mu ( \frac{1}{2} (\tanh ({\bf w}) + 1) - \bfx ) \; + \;
c \;  g( \frac{1}{2} (\tanh ({\bf w}) + 1))
\end{array}
\]
In the objective given above, the ``Lagrangian variable'' $c > 0$ is a
suitably chosen constant (from the optimization literature we know
that there exists $c > 0$ such that the optimal solutions of the last
two formulations are the same).

\subsection{Adversarial Training}
\label{subsec:adv-training}

Once an attacker finds an adversarial example, then the algorithm can
be retrained using this example. Researchers have found that
retraining the model with adversarial examples produces a more robust
model. For this section, we will work with attack algorithms that have
a target label $t$ (i.e. we are in the targeted mis-classification
case, such as JSMA or CW). Let $\mathcal{A}(w,\bfx,t)$ be the attack
algorithm, where its inputs are as follows: $w \in H$ is the current
hypothesis, $\bfx$ is the data point, and $t \in \calC$ is the target
label. The output of $\mathcal{A}(w,\bfx,t)$ is a perturbation
$\bfdelta$ such that $F(\bfx+\bfdelta) = t$. If the attack algorithm
is simply a mis-classification algorithm (e.g. FGSM or Deepfool) we
will drop the last parameter $t$. 

An {\it adversarial training} algorithm $\mathcal{R}_{\mathcal{A}}
(w,\bfx,t)$ is parameterized by an attack algorithm $\mathcal{A}$ and
outputs a new hypothesis $w' \in H$. Adversarial training works by
taking a datapoint $\bfx$ and an attack algorithm
$\mathcal{A}(w,\bfx,t)$ as its input and then retraining the
model using a specially designed loss function (essentially one
performs a single step of the SGD using the new loss function).  The
question arises: what loss function to use during the training?
Different methods use different loss functions.

Next, we discuss some adversarial training algorithms proposed in the
literature.  At a high level, an important point is that
the more sophisticated an adversarial perturbation algorithm is,
harder it is to turn it into adversarial training.  The reason is that
it is hard to ``encode'' the adversarial perturbation algorithm as an
objective function and optimize it. We will see this below, especially
for the virtual adversarial training (VAT) proposed by Miyato et
al.~\cite{miyato2015distributional}.

\noindent\textbf{Retraining for FGSM.}
We discussed the FGSM attack method earlier. 
In this case $\mathcal{A} \; = \; \FGSM$. The loss
function used by the retraining algorithm $\mathcal{R}_{\FGSM} (w,\bfx,t)$
is as follows:
\begin{align*}
  &\ell_{\FGSM}(w, \bfx_i, y_i)
  = \ell(w, \bfx_i, y_i)
    + \lambda \ell \left(w, \bfx_i + \FGSM(w,\bfx_i) ,y_i \right) \\
\end{align*}
Recall that $\FGSM (w,\bfx)$ was defined earlier, and $\lambda$ is a
regularization parameter.  The simplicity of $\FGSM(w,\bfx_i)$ allows
taking its gradient, but this objective function requires label $y_i$
because we are reusing the same loss function $\ell$ used to train the
original model.  Further, $ \FGSM(w,\bfx_i)$ may not be very good
because it may not produce good adversarial perturbation direction
(i.e. taking a bigger step in this direction might produce a distorted
image). The retraining algorithm is simply as follows: {\it take one
  step in the SGD using the loss function $\ell_{\FGSM}$ at the data
  point $\bfx_i$.}

A caveat is needed for taking gradient during the SGD step. At
iteration $t$ suppose we have model parameters $w_t$, and we need to
compute the gradient of the objective.  Note that $\FGSM(w,\bfx)$ depends
on $w$ so by chain rule we need to compute $\partial \FGSM(w,\bfx) /
\partial w |_{w=w_t}$.  However, this gradient is volatile~\footnote{In
  general, second-order derivatives of a classifier corresponding to a
  DNN vanish at several points because several layers are piece-wise
  linear.}, and so instead Goodfellow et al. only compute:
\begin{align*}
  \left.
  \frac{\partial \ell\left(w, \bfx_i + \FGSM(w_t, \bfx_i), 
  y_i \right)}{\partial w}
  \right|_{w=w_t}
\end{align*}
Essentially they treat $\FGSM (w_t, \bfx_i)$ as a constant while taking
the derivative.

\vskip 5pt
\noindent\textbf{Virtual Adversarial Training (VAT).}  Miyato et
al.~\cite{miyato2015distributional} observed the drawback of requiring
label $y_i$ for the adversarial example.  Their intuition is that one
wants the classifier to behave ``similarly'' on $\bfx$ and $\bfx +
\delta$, where $\delta$ is the adversarial perturbation. Specifically,
the distance of the distribution corresponding to the output of the softmax layer
$F_w$ on $\bfx$ and $\bfx + \delta$ is small.  VAT uses {\it
  Kullback–Leibler ($\KL$) divergence} as the measure of the distance between
two distributions.  Recall that $\KL$ divergence of two distributions
$P$ and $Q$ over the same finite domain $D$ is given by the following
equation:
\begin{eqnarray*}
\KL (P,Q) & = & \sum_{i \in D} P(i) \log \left( \frac{P(i)}{Q(i)} \right) 
\end{eqnarray*}

Therefore, they propose that, instead of reusing $\ell$, they propose
to use the following for the regularizer,
\begin{align*}
  \Delta(r, \bfx, w) =
  \KL\left(s(F_w)(\bfx)[y], s(F_w)(\bfx+r)[y] \right)
\end{align*}
for some $r$ such that $\|r\| \le \delta$.  As a result, the label $y_i$ is
\emph{no longer} required. The question is: what $r$ to use?  Miyato
et al.~\cite{miyato2015distributional} propose that in theory we
should use the ``best'' one as
\begin{align*}
  \max_{r:\|r\| \le \delta} \; \; \KL\left(s(F_w)(\bfx)[y], s(F_w)(\bfx+r)[y] \right)
\end{align*}

This thus gives rise to the following loss function to use during retraining:
\begin{align*}
  &\ell_{\VAT}(w, \bfx_i, y_i)
  = \ell(w, , \bfx_i,y_i)
    + \lambda 
      \max_{r:\|r\| \le \delta}\Delta(r, \bfx_i, w)
\end{align*}
However, one cannot easily compute the gradient for the regularizer.
Hence the authors perform an  approximation as follows:
\begin{enumerate}
\item Compute the Taylor expansion of $\Delta(r, \bfx_i, w)$ at $r = 0$, so
  $\Delta(r, \bfx_i, w) = r^T H(\bfx_i, w) \; r$ where $H(\bfx_i, w)$
  is the Hessian matrix of $\Delta(r, \bfx_i, w)$ with respect to $r$ at $r = 0$.

\item Thus $\max_{\|r\| \le \delta}\Delta(r, \bfx_i, w)
  = \max_{\|r\| \le \delta}\left(r^T H(\bfx_i, w) \; r \right)$.
  By variational characterization of the symmetric matrix ($H(\bfx_i, w)$ is symmetric),
  $r^* = \delta \bar{v}$ where $\bar{v} = \overline{v(\bfx_i, w)}$ is the unit
  eigenvector of $H(\bfx_i, w)$ corresponding to its largest eigenvalue.
  Note that $r^*$ depends on $\bfx_i$ and $w$. Therefore the loss function becomes:
  \begin{align*}
    \ell_{\VAT}(\theta, \bfx_i, y_i)
  = \ell(\theta, \bfx_i, y_i) + \lambda \Delta(r^*, \bfx_i, w)
  \end{align*}

\item Now suppose in the process of SGD we are at iteration $t$ with model parameters $w_t$,
  and we need to compute $\partial \ell_{\VAT} / \partial w |_{w=w_t}$.
  By chain rule we need to compute
  $\partial r^* / \partial w |_{w=w_t}$.
  However the authors find that such gradients are volatile, so they instead fix
  $r^*$ as a constant at the point $\theta_t$, and compute
  \begin{align*}
    \left.\frac{\partial \KL\left(s(F_w)(\bfx)[y], s(F_w)(\bfx+r)[y] \right)}
    {\partial w}\right|_{w=w_t}
  \end{align*}
  
\end{enumerate}


\subsection{Black Box Attacks}
\label{subsec:black-box}

Recall that earlier attacks (e.g. FGSM and JSMA) needed white-box
access to the classifier $F$ (essentially because these attacks
require first order information about the classifier). In this
section, we present black-box attacks.  In this case, an attacker can
{\it only} ask for the labels $F(\bfx)$ for certain data points. Our
presentation is based on~\cite{papernot:BB}, but is
more general.

Let $\mathcal{A}(w,\bfx,t)$ be the attack algorithm, where its inputs
are: $w \in H$ is the current hypothesis, $\bfx$ is the data point,
and $t \in \calC$ is the target label. The output of
$\mathcal{A}(w,\bfx,t)$ is a perturbation $\bfdelta$ such that
$F(\bfx+\bfdelta) = t$. If the attack algorithm is simply a
mis-classification algorithm (e.g. FGSM or Deepfool) we will drop the
last parameter $t$ (recall that in this case the attack algorithm
returns a $\bfdelta$ such that $F(\bfx+\bfdelta) \not= F(\bfx)$). An
{\it adversarial training} algorithm $\mathcal{R}_{\mathcal{A}}
(w,\bfx,t)$ is parameterized by an attack algorithm $\mathcal{A}$ and
outputs a new hypothesis $w' \in H$ (this was discussed in the
previous subsection).

\noindent
{\it Initialization:} We pick a substitute classifier $G$ and an initial
seed data set $S_0$ and train $G$. For simplicity, we will assume that
the sample space $Z = X \times Y$ and the hypothesis space $H$ for $G$
is same as that of $F$ (the classifier under attack). However, this is
not crucial to the algorithm. We will call $G$ the {\it substitute
  classifier} and $F$ the {\it target classifier}. Let $S=S_0$ be the initial
data set, which will be updated as we iterate.

\noindent
{\it Iteration:} Run the attack algorithm $\mathcal{A}(w,\bfx,t)$ on $G$
and obtain a $\bfdelta$. If $F(\bfx+\bfdelta) = t$, then {\bf stop} we are
done. If $F(\bfx+\bfdelta) = t'$ but not equal to $t$, we augment the data
set $S$ as follows:
\begin{eqnarray*}
S & = & S \cup { (\bfx+\bfdelta, t')}
\end{eqnarray*}
We now retrain $G$ on this new data set, which essentially means
running the SGD on the new data point $(\bfx+\bfdelta, t')$. Notice
that we can also use adversarial training $\mathcal{R}_{\mathcal{A}}
(w,\bfx,t)$ to update $G$ (to our knowledge this has been not tried
out in the literature).

\subsection{Defenses}

Defenses with formal guarantees against test-time attacks have proven
elusive. For example, Carlini and Wagner~\cite{Carlini-Wagner:2017}
have a recent paper that breaks {\it ten recent defense
proposals}. However, defenses that are based on robust-optimization
objectives have demonstrated promise~\cite{Madry:2018,Kolter,Duchi}.
Several techniques for verifying properties of a DNN (in isolation) have appeared
recently (e.g.,~\cite{Barrett:17,Kwiatkowska:17,dutta-nfm18,dvijotham-arxiv18}).  
Due to space limitations we will not give a detailed account of all these defenses.


\section{Semantic Adversarial Analysis and Training}
\label{sec:semantic-analysis}

A central tenet of this paper is that the analysis of deep
neural networks (and machine learning components, in general)
must be more {\em semantic}. 
In particular, we advocate for the increased use of semantics
in several aspects of adversarial analysis and training,
including the following:
\begin{myitemize}
\item
{\em Semantic Modification Space:}
Recall that the goal of adversarial attacks is to modify an
input vector $\bfx$ with an adversarial modification $\bfdelta$
so as to achieve a target misclassification. Such modifications
typically do not incorporate the application-level semantics or
the context within which the neural network is deployed.
We argue that it is essential to incorporate more application-level,
contextual semantics into the modification space. Such
{\em semantic modifications} correspond to modifications that may
arise more naturally within the context of the target application.
We view this not as ignoring arbitrary modifications (which are
indeed worth considering with a security mind set), but as prioritizing
the design and analysis of DNNs towards semantic adversarial
modifications. Sec.~\ref{sec:comp-falsif} discusses this point in
more detail.

\item
{\em System-Level Specifications:}
The goal of much of the work in adversarial attacks has been to
generate misclassifications. However, not all misclassifications are
made equal. We contend that it is important to find misclassifications
that lead to violations of desired properties of the system within
which the DNN is used. Therefore, one must identify such {\em system-level
specifications} and devise analysis methods to verify whether 
an erroneous behavior of the DNN component can lead to the violation
of a system-level specification. System-level counterexamples can
be valuable aids to repair and re-design machine learning models.
See Sec.~\ref{sec:comp-falsif} for a more detailed discussion of
this point.

\item
{\em Semantic (Re-)Training:}
Most machine learning models are trained with the main goal of
reducing misclassifications as measured by a suitably crafted
loss function. We contend that it is also important to train
the model to avoid undesirable behaviors at the system level.
For this, we advocate using methods for {\em semantic training},
where system-level specifications, counterexamples, and other
artifacts are used to improve the semantic quality of the ML model.
Sec.~\ref{sec:sem-train} explores a few ideas.

\item
{\em Confidence-Based Analysis and Decision Making:}
Deep neural networks (and other ML models) often produce not
just an output label, but also an associated confidence level.
We argue that {\em confidence levels} must be used within the design
of ML-based systems. They provide a way of exposing more information
from the DNN to the surrounding system that uses its decisions.
Such confidence levels can also be useful to prioritize analysis
towards cases that are more egregious failures of the DNN.
More generally, any {\em explanations} and {\em auxiliary information}
generated by the DNN that accompany its main output decisions 
can be valuable aids in their design and analysis.

\end{myitemize}

\subsection{Compositional Falsification}
\label{sec:comp-falsif}

We discuss the problem of performing system-level analysis of a deep
learning component, using recent work by the
authors~\cite{dreossi-nfm17,dreossi-arxiv17} to illustrate the main
points. The material in this section is mainly based
on~\cite{seshia-tr17}.

We begin with some basic notation.
Let $S$ denote the model of the full system $S$ under verification,
$E$ denote a model of its environment,
and $\Phi$ denote the specification to be verified.
$C$ is an ML model (e.g. DNN) that is part of $S$.
As in Sec.~\ref{sec:attacks}, let $\bfx$ be an input to $C$.
We assume that $\Phi$ is a trace property -- a set of behaviors of the closed system
obtained by composing $S$ with $E$, denoted $S \| E$.
The goal of falsification is to find one or more counterexamples
showing how the composite system $S \| E$ violates
$\Phi$. In this context, {\em semantic analysis of $C$ is about
finding a modification $\bfdelta$ from a space of semantic
modifications $\Delta$ such that $C$, on $\bfx+\bfdelta$,
produces a misclassification that causes $S \| E$ to violate $\Phi$}.

\subsubsection{Example Problem}
\label{sec:example}

\begin{figure}
  	\begin{center}
	\includegraphics[width=0.8\textwidth]{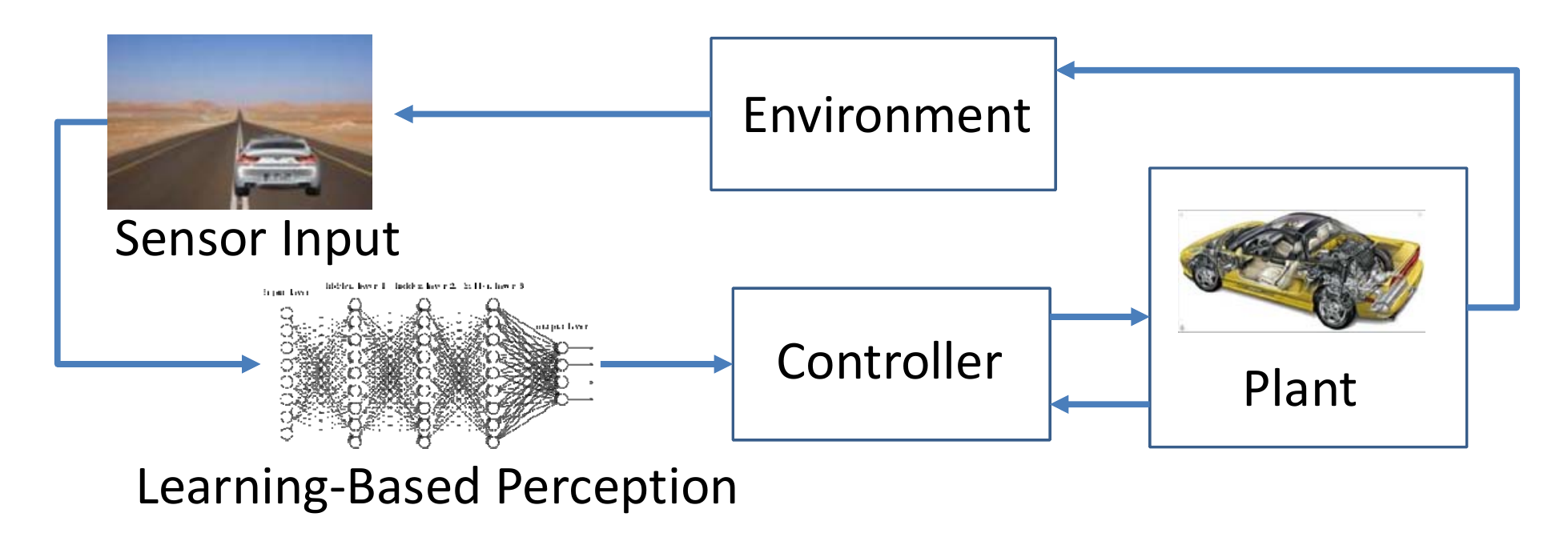}
	 \end{center}
	\caption{Automatic Emergency Braking System (AEBS) in closed loop. An image classifier based on deep neural networks is used to perceive objects in the ego vehicle's frame of view.\label{fig:aebs}}
\end{figure}

As an illustrative example, consider a simple model of an Automatic Emergency Braking System (AEBS),
that attempts to detect objects in front of a vehicle and actuate
the brakes when needed to avert a collision.
Figure~\ref{fig:aebs} shows the AEBS as a system
composed of a controller (automatic braking), a plant (vehicle sub-system under control, including transmission), and an advanced sensor (camera along
with an obstacle detector based on deep learning). The AEBS, when combined
with the vehicle's environment, forms a closed loop control system.
The controller regulates the acceleration and braking of the plant using the velocity of
the subject (ego) vehicle and the distance between it and an obstacle.
The sensor used to detect the obstacle includes a camera along with an image classifier based on DNNs.
In general, this sensor can provide noisy measurements due to incorrect image 
classifications which in turn can affect the correctness of the overall system.

Suppose we want to verify 
whether the distance between the ego vehicle and a preceding obstacle is always larger than
$2$ meters. In STL, this requirement $\Phi$ can be written as
$\G{0,T} (\|\mathbf{x}_{\text{ego}} - \mathbf{x}_{\text{obs}}\|_2 \geq 2)$.
Such verification requires the exploration of a very 
large input space comprising of
the control inputs (e.g., acceleration and braking pedal angles) and 
the machine learning (ML) component's feature space
(e.g., all the possible pictures observable by the camera). 
The latter space is particularly large --- for example,
note that the feature space of RGB images of dimension 
$1000\times600$px (for an image classifier) 
contains $256^{1000\times 600 \times 3}$ elements.

In the above example, $S \| E$ is the closed loop system in
Fig.~\ref{fig:aebs} where $S$ comprises the DNN and the controller,
and $E$ comprises everything else.  $C$ is the DNN used for object
detection and classification.

This case study has been implemented in
Matlab/Simulink\footnote{\url{https://github.com/dreossi/analyzeNN}}
in two versions that use two different Convolutional Neural Networks
(CNNs): the Caffe~\cite{jia2014caffe} version of
AlexNet~\cite{krizhevsky2012imagenet} and the Inception-v3 model
created with Tensorflow~\cite{tensorflow2015}, both trained on the
ImageNet database~\cite{imagenet}.  Further details about this example
can be obtained from~\cite{dreossi-nfm17}.

\subsubsection{Approach}

\begin{figure}
\begin{center}
\includegraphics[width=\textwidth]{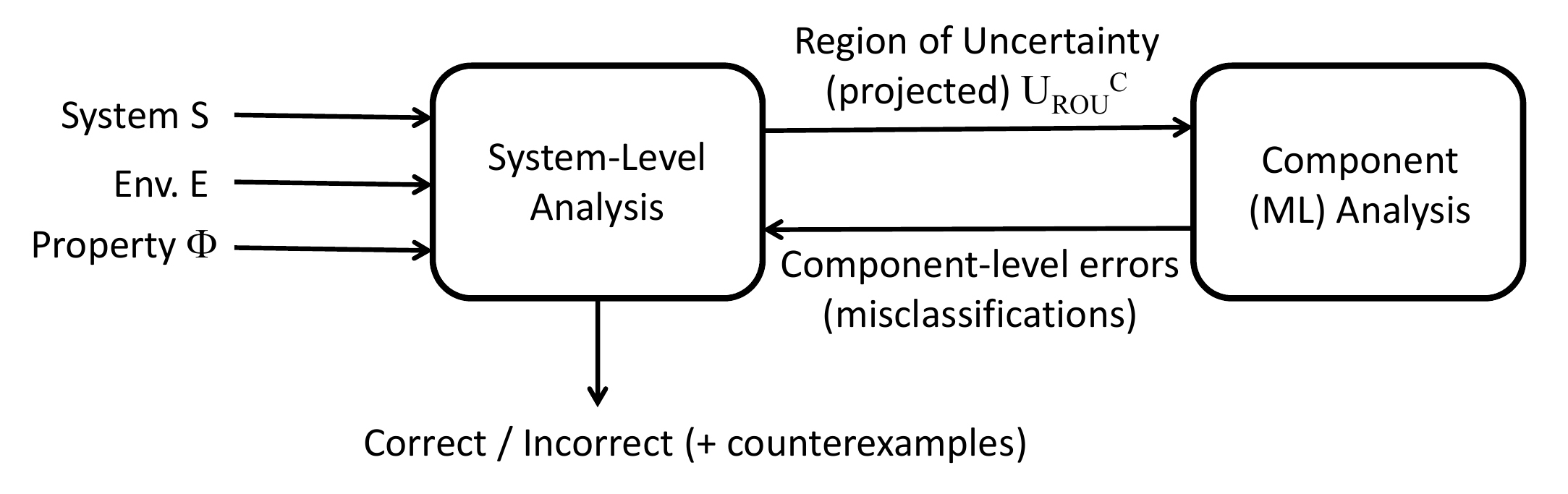}
\end{center}
\caption{Compositional Verification Approach. A system-level verifier cooperates with a component-level analysis procedure (e.g., adversarial analysis of a machine learning component to find misclassifications). \label{fig:CompApproach}}
\end{figure}

A key idea in our approach is to have a {\em system-level verifier}
that abstracts away the component
$C$ while verifying $\Phi$ on the resulting abstraction.
This system-level verifier communicates with a component-level
analyzer that searches for semantic modifications $\bfdelta$ 
to the input $\bfx$ of $C$ 
that could lead to violations of the system-level specification
$\Phi$.
Figure~\ref{fig:CompApproach} illustrates this approach.

We formalize this approach while trying to 
emphasize the intuition. 
Let $T$ denote the set of all possible traces of the composition
of the system with its environment, $S \| E$.
Given a specification $\Phi$, let $T_{\Phi}$ denote the
set of traces in $T$ satisfying $\Phi$. Let $U_{\Phi}$ denote
the projection of these traces onto the state and interface
variables of the environment $E$.
$U_{\Phi}$ is termed as the {\em validity domain} of $\Phi$,
i.e., the set of environment behaviors for which $\Phi$ is
satisfied. Similarly, the complement set $U_{\neg\Phi}$ is
the set of environment behaviors for which $\Phi$ is violated.

\begin{figure}[ht]
 \centering
 \includegraphics[width=\textwidth]{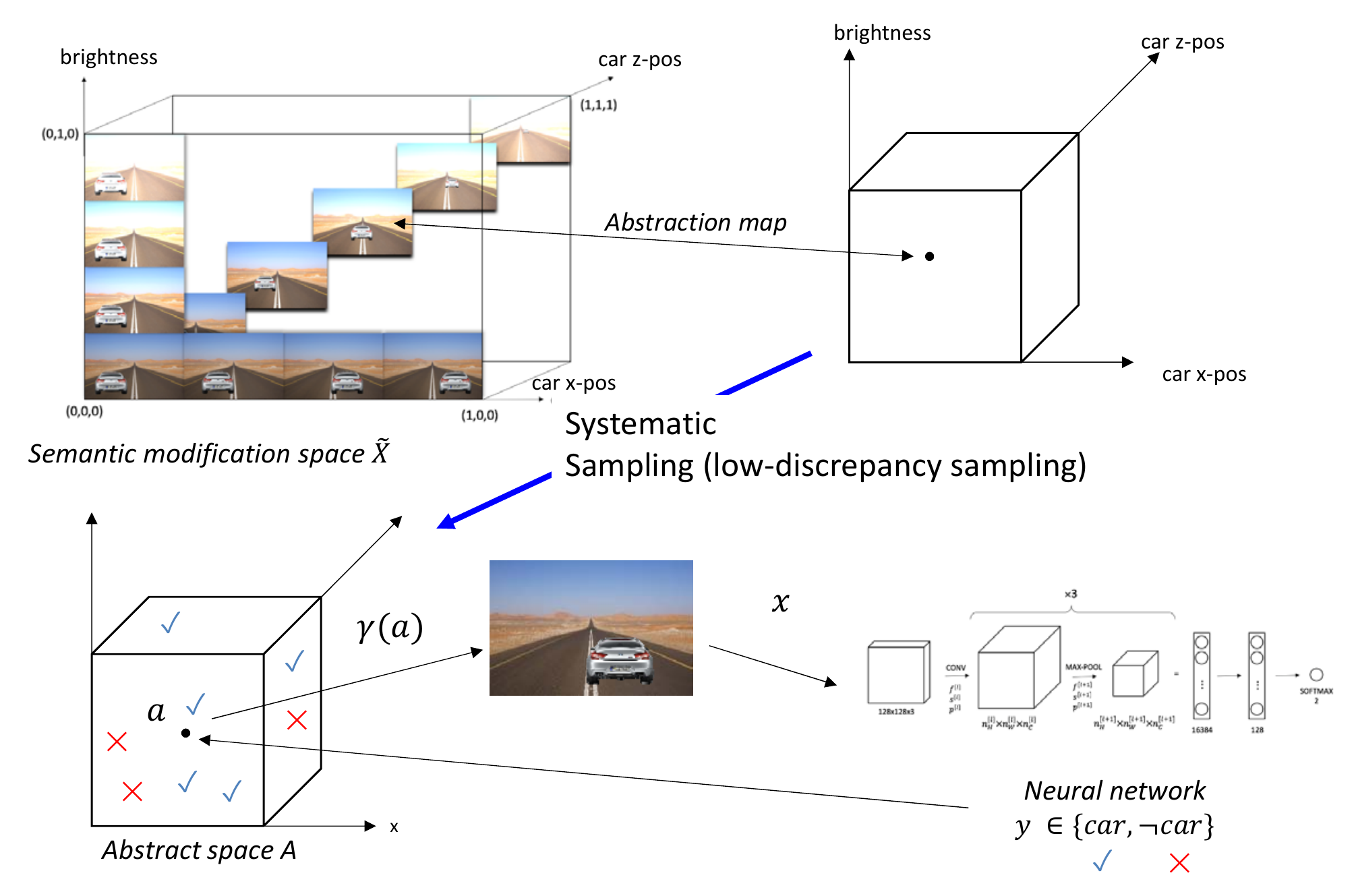}
	\caption{Machine Learning Analyzer: Searching the Semantic Modification Space. {\small{A concrete semantic modification space (top left) is mapped into a discrete abstract space. Systematic sampling, using low-discrepancy methods, yields points in the abstract space. These points are concretized and the NN is evaluated on them to ascertain if they are correctly or wrongly classified. The misclassifications are fed back for system-level analysis.}}}
\label{fig:ml-analyzer-flow}
\end{figure}

Our approach works as follows:
\begin{enumerate}
\item The System-level Verifier initially performs two analyses
with two extreme abstractions of the ML component. First, it
performs an {\em optimistic} analysis, wherein the ML component is
assumed to be a ``perfect classifier'', i.e.,
all feature vectors are correctly classified. In situations where
ML is used for perception/sensing, this abstraction assumes perfect
perception/sensing. Using this abstraction,
we compute the validity domain for this abstract model of the system,
denoted $U^+_{\Phi}$. 
Next, it performs a {\em pessimistic} analysis where the ML
component is abstracted by a ``completely-wrong classifier'',
i.e., all feature vectors are misclassified. Denote the resulting
validity domain as $U^-_{\Phi}$. It is expected that 
$U^+_{\Phi} \supseteq U^-_{\Phi}$. 

Abstraction permits the System-level Verifier to operate on a 
lower-dimensional search space and identify a region in this space 
that may be affected by the malfunctioning of component $C$ ---
 a so-called ``region of uncertainty'' (ROU).
This region, $U^C_{ROU}$ is computed as $U^+_{\Phi} \setminus U^-_{\Phi}$.
In other words, it comprises all environment behaviors that could
lead to a system-level failure when component $C$ malfunctions.
This region $U^C_{ROU}$, projected onto the inputs of $C$, 
is communicated to the ML Analyzer.
(Concretely, in the context of our example of Sec.~\ref{sec:example},
this corresponds to finding a subspace of images that corresponds to
$U^C_{ROU}$.)

\item The Component-level Analyzer, also termed as a Machine Learning 
(ML) Analyzer, performs a detailed analysis of the projected ROU
$U^C_{ROU}$. 
A key aspect of the ML analyzer is to explore the
{\em semantic modification space} efficiently.
Several options are available for such an analysis,
including the various adversarial analysis techniques surveyed earlier
(applied to the semantic space),
as well as systematic sampling methods~\cite{dreossi-nfm17}.
Even though a component-level formal specification may not be available,
each of these adversarial analyses has an implicit notion of
``misclassification.'' We will refer to these as {\em component-level errors}. The working of the ML analyzer from~\cite{dreossi-nfm17} is
shown in Fig.~\ref{fig:ml-analyzer-flow}.

\item When the Component-level (ML) Analyzer finds 
component-level errors (e.g., those that trigger misclassifications 
of inputs whose labels are easily inferred), it communicates that 
information back to the System-level Verifier, which checks whether 
the ML misclassification can lead to a violation of the 
system-level property $\Phi$. If yes, we have found a system-level 
counterexample. If no component-level errors are found, and the
system-level verification can prove the absence of counterexamples,
then it can conclude that $\Phi$ is satisfied.
Otherwise, if the ML misclassification cannot be extended to
a system-level counterexample, the ROU is updated and the 
revised ROU passed back to the Component-level Analyzer. 

\end{enumerate}
The communication between the System-level Verifier and the
Component-level (ML) Analyzer continues thus, until we either
prove/disprove $\Phi$, or we run out of resources.

\subsubsection{Sample Results}

We have applied the above approach to the problem of 
{\em compositional falsification} of cyber-physical systems (CPS)
with machine learning components~\cite{dreossi-nfm17}. For this class of CPS,
including those with highly non-linear dynamics and even black-box 
components, simulation-based falsification of temporal logic properties
is an approach that has proven effective in industrial practice
(e.g.,~\cite{jin-tcad15,yamaguchi-fmcad16}).
We present here a sample of results on the AEBS example from~\cite{dreossi-nfm17},
referring the reader to more detailed descriptions in the
other papers on the topic~\cite{dreossi-nfm17,dreossi-arxiv17}.

\begin{figure}[h]
\begin{center}
\includegraphics[scale=0.4]{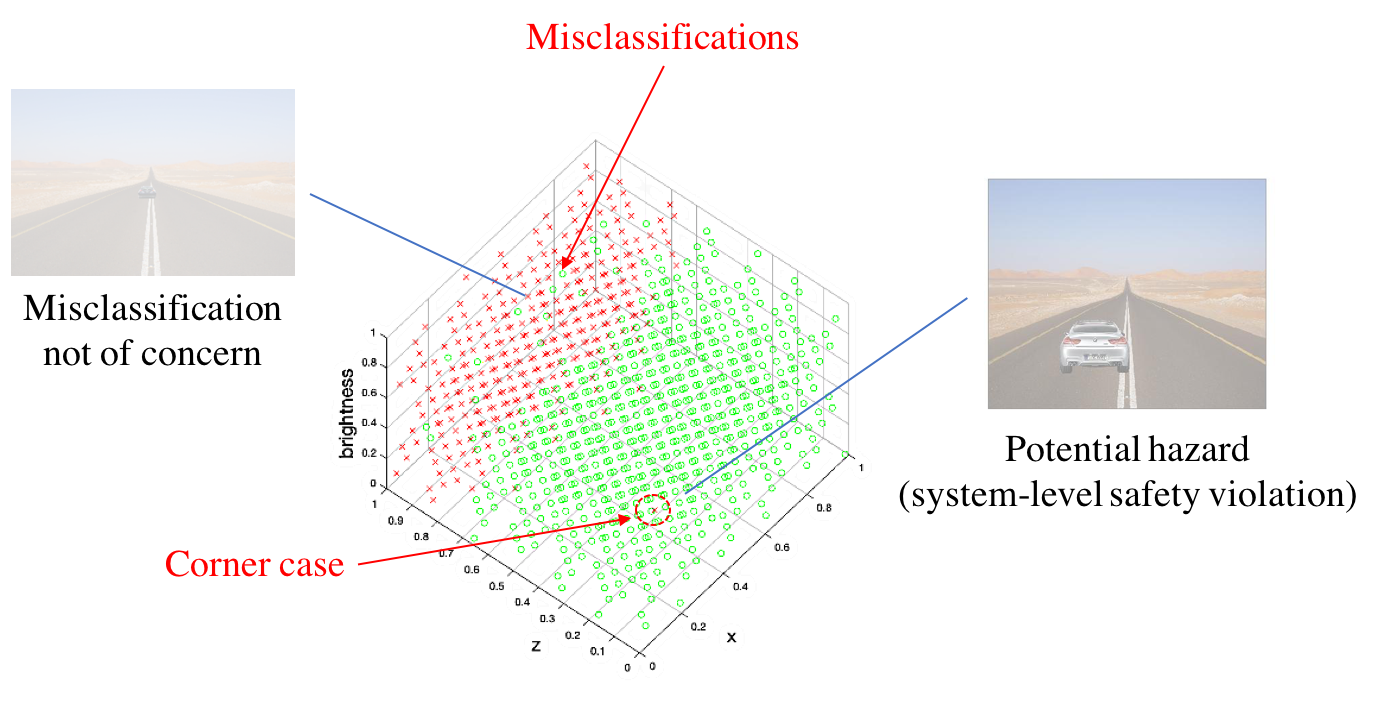}
\end{center}
\caption{Misclassified Images for Inception-v3 Neural Network (trained on ImageNet with TensorFlow). Red crosses are misclassified images and green circles are correctly classified. Our system-level analysis finds a corner-case image that could lead to a system-level safety violation. \label{fig:InceptionErrors}}
\end{figure}

%
%
In Figure~\ref{fig:InceptionErrors} we show one result of our analysis
for the Inception-v3 deep neural network. This figure shows both
correctly classified and misclassified images on a range of synthesized
images where (i) the environment vehicle is moved away from or towards
the ego vehicle (along z-axis), 
(ii) it is moved sideways along the road (along x-axis),
or (iii) the brightness of the image is modified. These modifications
constitute the 3 axes of the figure. Our approach finds misclassifications
that do not lead to system-level property violations and also
misclassifications that do lead to such violations.
For example, Figure~\ref{fig:InceptionErrors} shows two 
misclassified images, one with an environment vehicle that
is too far away to be a safety hazard, as well as another
image showing an environment vehicle driving slightly on the wrong
side of the road, which is close enough to potentially cause
a violation of the system-level safety property (of maintaining
a safe distance from the ego vehicle).

For further details about this and other results with our approach,
we refer the reader to~\cite{dreossi-nfm17,dreossi-arxiv17}.

\subsection{Semantic Training}
\label{sec:sem-train}

In this section we discuss two ideas for {\em semantic training and retraining}
of deep neural networks. 
We first discuss the use of {\em hinge loss} as
a way of incorporating confidence levels into the training process.
Next, we discuss how system-level counterexamples and associated misclassifications
can be used in the retraining process to both improve the accuracy of
ML models and also to gain more assurance 
in the overall system containing the ML component.
A more detailed study of using misclassifications 
(ML component-level counterexamples)
to improve the accuracy of the neural network 
is presented in~\cite{dreossi-ijcai18};
this approach is termed {\em counterexample-guided data augmentation},
inspired by counterexample-guided abstraction refinement
(CEGAR)~\cite{clarke-cav00} and similar paradigms.

\subsubsection{Experimental Setup}

As in the preceding section,
we consider an Automatic Emergency Braking System (AEBS) using a DNN-based
object detector. 
However, in these experiments we use an AEBS deployed within Udacity's 
self-driving car simulator, as reported in our previous work~\cite{dreossi-arxiv17}.\footnote{Udacity's self-driving car simulator: \url{https://github.com/udacity/self-driving-car-sim}}
We modified the Udacity simulator to focus exclusively on braking. 
In our case studies, the car follows some predefined way-points,
while accelerating and braking are controlled by the AEBS connected to a
convolutional neural network (CNN). In particular, whenever the
CNN detects an obstacle in the images provided by the onboard camera, 
the AEBS triggers a braking action
that slows the vehicle down and avoids the collision against the obstacle.

\begin{figure}
	\centering
	\includegraphics[scale=0.3]{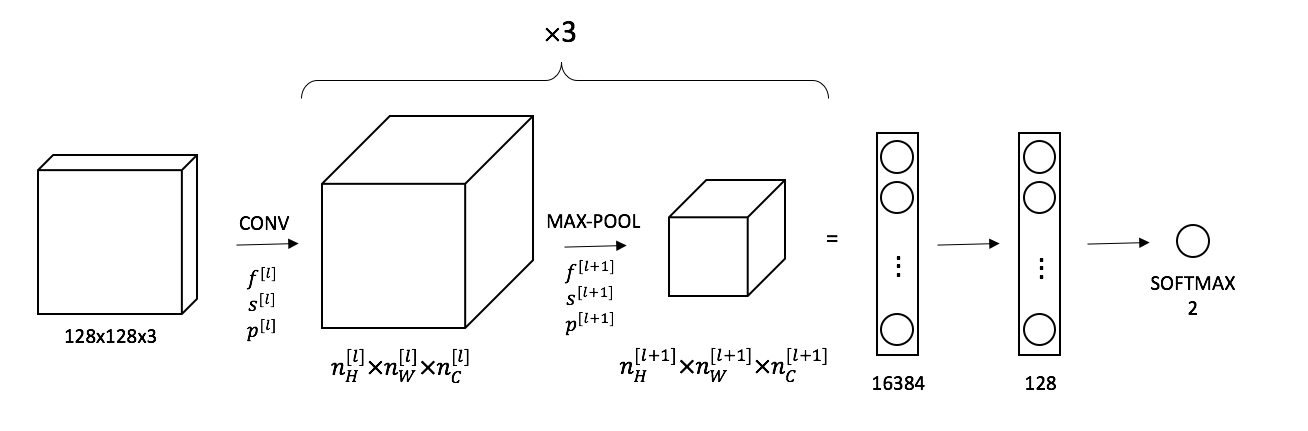}
	\caption{CNN architecture.\label{fig:cnn_archi}}
\end{figure}

We designed and implemented a CNN to predict the presence of a cow on the road.
Given an image taken by the onboard camera, the CNN classifies the picture in either \lq\lq cow\rq\rq\
or \lq\lq not cow\rq\rq\ category.
The CNN  architecture  is  shown  in  Fig.~\ref{fig:cnn_archi}.  It consists of
eight layers: the first six are alternations of convolutions and max-pools with
ReLU activations, the last two are a fully connected layer and a softmax that
outputs the network prediction (confidence level for each label). 

\begin{figure}
	\centering
	\includegraphics[scale=0.5]{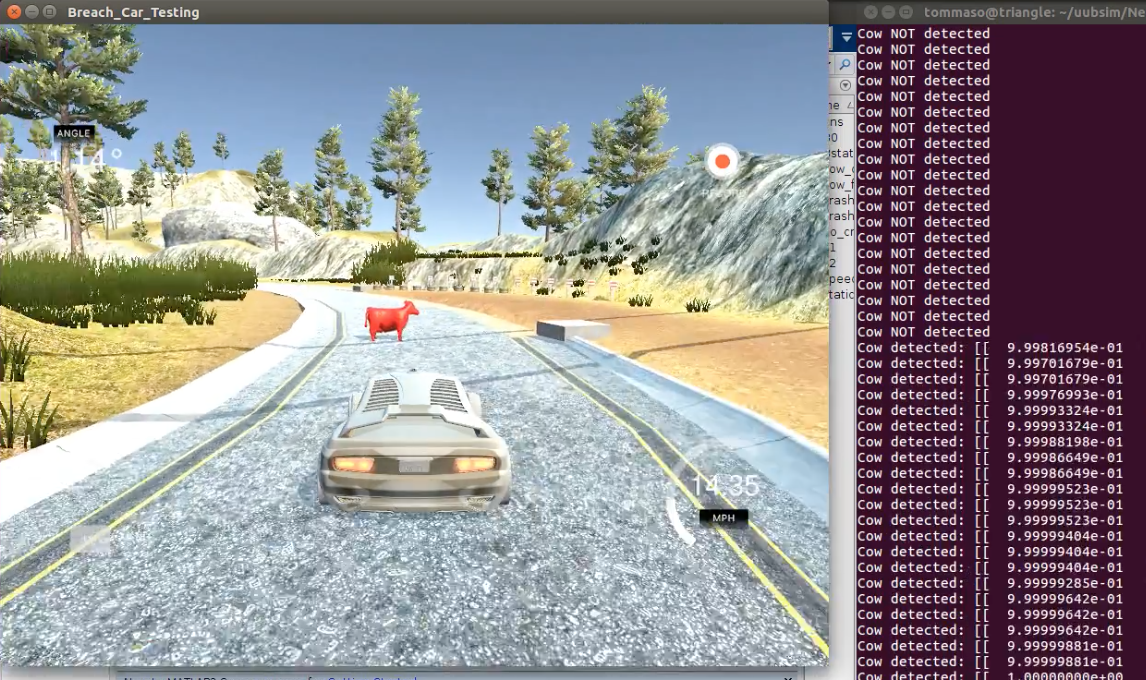}
	\caption{Udacity simulator with a CNN-based AEBS in action.\label{fig:aebs_cow}}
\end{figure}

We generated a data set of 1000 road images with and without cows.  We
split the data set into $80\%$ training and $20\%$ validation
data. Our model was implemented and trained using the Tensorflow
library with cross-entropy cost function and the Adam algorithm
optimizer (learning rate $10^{-4}$). The model reached $95\%$ accuracy
on the test set.  Finally, the resulting CNN is connected to the Unity
simulator via Socket.IO protocol.\footnote{Socket.IO
protocol: \url{https://github.com/socketio}} Fig.~\ref{fig:aebs_cow}
depicts a screenshot of the simulator with the AEBS in action in
proximity of a cow.

\subsubsection{Hinge Loss}\label{sec:hinge_loss}

In this section, we investigate the relationship between multiclass
hinge loss functions and adversarial examples. {\it Hinge loss} is defined
as follows:
\begin{equation}
	l(\hat{y}) = \max(0, k+ \max_{i \neq l}(\hat{y}_i) - \hat{y}_l )
\end{equation}
where $(x,y)$ is a training sample, $\hat{y} = F(x)$ is a prediction,
and $l$ is the {\it ground truth} label of $x$.
For this section, the output $\hat{y}$ is a numerical value indicating
the {\em confidence level} of the network for each class. For example,
$\hat{y}$ can be the output of a softmax layer as described in 
Sec.~\ref{sec:background}. 

Consider what happens as we vary $k$. Suppose there is an $i \neq l$
s.t. $\hat{y_i} > \hat{y}_l$. Pick the largest such $i$, call it $i^*$.
For $k=0$, we will incur a loss of $\hat{y}_{i^*} - \hat{y}_l$ for the
example $(x,y)$. However, as we make $k$ more negative, we increase
the tolerance for ``misclassifications'' produced by the DNN $F$.
Specifically, we incur no penalty for a misclassification 
as long as the associated confidence level deviates from that of the 
ground truth label by no more than $|k|$. Larger the absolute value
of $k$, the greater the tolerance.
Intuitively, this biases the training process towards avoiding
``high confidence misclassifications''. 

In this experiment, we investigate the role of $k$
and explore different parameter values. At training time,
we want to minimize the mean hinge loss across all training samples.
We trained the CNN described above 
with different values of $k$ and evaluated its precision on
both the original test set and a set of counterexamples generated
for the original model, i.e., the network trained with cross-entropy loss.

Table~\ref{tab:hinge} reports accuracy and log loss for different
values of $k$ on both original and counterexamples test sets
($T_{original}$ and $T_{countex}$, respectively).

\begin{table}
	\centering
	\begin{tabular}{c | c c | c c}
		& \multicolumn{2}{c}{$T_{original}$} & \multicolumn{2}{| c}{$T_{countex}$}\\
		\hline
		$k$ & acc & log-loss & acc & log-loss\\
		\hline
		0 & 0.69 & 0.68 & 0.11 & 0.70 \\
		- 0.01 & 0.77 & 0.69 & 0.00 & 0.70\\
		-0.05 & 0.52 & 0.70 & 0.67 & 0.69\\
		-0.1 & 0.50 & 0.70 & 0.89 & 0.68\\
		-0.25 & 0.51 & 0.70 & 0.77 & 0.68\\
	\end{tabular}
	\caption{Hinge loss with different $k$ values.\label{tab:hinge}}
\end{table}

Table~\ref{tab:hinge} shows interesting results. We note that
a negative $k$ increases the accuracy of the model on counterexamples.
In other words, biasing the training process by penalizing 
high-confidence misclassifications improves accuracy on
counterexamples!
However, the price to pay is a reduction of accuracy on the original test set.
This is still a very preliminary result and further experimentation
and analysis is necessary.

\subsubsection{System-Level Counterexamples}

\begin{figure}
	\centering
	\includegraphics[scale=0.5]{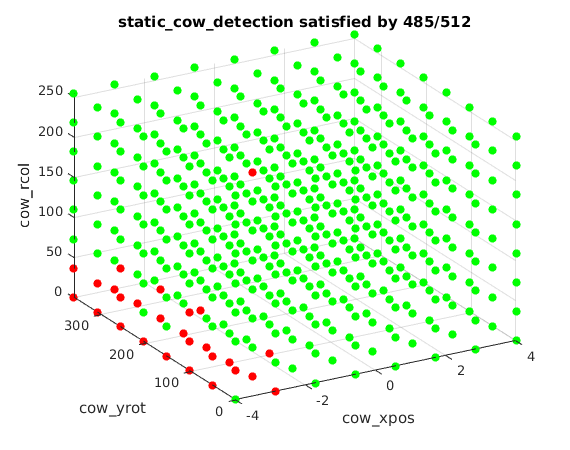}
	\caption{Semantic counterexamples: obstacle configurations leading to property violations (in red).\label{fig:cow_analysis}}
\end{figure}

By using the composition falsification framework presented in
Sec.~\ref{sec:comp-falsif}, we identify orientations, displacements on
the $x$-axis, and color of an obstacle that leads to a collision of
the vehicle with the obstacle. Figure~\ref{fig:cow_analysis} depicts
configurations of the obstacle that lead to specification violations,
and hence, to collisions.

In an experiment, we augment
the original training set with the elements of
$T_{countex}$, i.e., images of the original test set $T_{original}$
that are misclassified by the original model (see Sec.~\ref{sec:hinge_loss}).

We trained the model with both cross-entropy and hinge loss for $20$ epochs.
Both models achieve a high accuracy on the validation set ($\approx 92\%$).
However, when plugged into the AEBS, neither of these models
prevents the vehicle from colliding
against the obstacle with an adversarial configuration.
This seems to indicate that simply retraining with some semantic (system-level) counterexamples
generated by analyzing the system containing the ML model
may not be sufficient to eliminate all semantic counterexamples.

Interestingly, though, it appears that in both cases the impact of the vehicle with the
obstacle happens at a slower speed than the one with the original model. In other
words, the AEBS system starts detecting the obstacle earlier than with the original model, 
and therefore starts braking earlier as well.
This means that despite the specification violations, the counterexample
retraining procedure seems to help with limiting the damage in case of a collision.
Coupled with a run-time assurance framework (see~\cite{seshia-arxiv16}), semantic
retraining could help mitigate the impact of misclassifications on the system-level
behavior.



\section{Conclusion}
\label{sec:concl}
In this paper, we surveyed the field of adversarial machine learning
with a special focus on deep learning and on test-time attacks. 
We then introduced the idea of {\em semantic adversarial
machine (deep) learning}, where adversarial analysis and
training of ML models is performed using the
semantics and context of the overall system within which the
ML models are utilized. We identified several ideas for integrating
semantics into adversarial learning, including using a semantic
modification space, system-level formal specifications, 
training using semantic counterexamples, and utilizing
more detailed information about the outputs produced by the 
ML model, including confidence levels, in the modules that use
these outputs to make decisions.
Preliminary experiments show the promise of these ideas, but
also indicate that much remains to be done.
We believe the field of semantic adversarial learning will be
a rich domain for research at the intersection of machine
learning, formal methods, and related areas.

\subsection*{Acknowledgments}
The first and third author were supported in part by
NSF grant 1646208, the DARPA BRASS program under
agreement number FA8750-16-C0043, the DARPA Assured Autonomy
program, and Berkeley Deep Drive.

\newpage

\bibliographystyle{plain}
\bibliography{somesh,xi,paper,biblio_tom}

\end{document}